\documentclass[letterpaper, 10 pt, conference]{ieeeconf}  

\IEEEoverridecommandlockouts                              

\usepackage{graphicx} 
\usepackage{booktabs} 
\usepackage{cite}
\usepackage{amsmath} 

\usepackage{booktabs}
\usepackage{amssymb} 
\usepackage{color}
\usepackage{colortbl}
\usepackage[colorlinks,linkcolor=blue,anchorcolor=blue,citecolor=blue]{hyperref}
\usepackage{cleveref}

\title{\LARGE \bf
U-DiT Policy: U-shaped Diffusion Transformers for Robotic Manipulation
}

\author{Linzhi Wu, Aoran Mei, Xiyue Wang, Guo-Niu Zhu*, ~\IEEEmembership{Member,~IEEE} and Zhongxue Gan*
\thanks{*Corresponding author: Guo-Niu Zhu and Zhongxue Gan.}
\thanks{All authors are with the College of Intelligent Robotics and Advanced Manufacturing, Fudan University, Shanghai 200433, China (e-mail: guoniu\_zhu@fudan.edu.cn; ganzhongxue@fudan.edu.cn).}
}

\begin{document}

\maketitle
\thispagestyle{empty}
\pagestyle{empty}

\begin{abstract}
Diffusion-based methods have been acknowledged as a powerful paradigm for end-to-end visuomotor control in robotics. Most existing approaches adopt a Diffusion Policy in U-Net architecture (DP-U), which, while effective, suffers from limited global context modeling and over-smoothing artifacts. To address these issues, we propose U-DiT Policy, a novel U-shaped Diffusion Transformer framework. U-DiT preserves the multi-scale feature fusion advantages of U-Net while integrating the global context modeling capability of Transformers, thereby enhancing representational power and policy expressiveness. We evaluate U-DiT extensively across both simulation and real-world robotic manipulation tasks. In simulation, U-DiT achieves an average performance gain of 10\% over baseline methods and surpasses Transformer-based diffusion policies (DP-T) that use AdaLN blocks by 6\% under comparable parameter budgets. On real-world robotic tasks, U-DiT demonstrates superior generalization and robustness, achieving an average improvement of 22.5\% over DP-U. In addition, robustness and generalization experiments under distractor and lighting variations further highlight the advantages of U-DiT. These results highlight the effectiveness and practical potential of U-DiT Policy as a new foundation for diffusion-based robotic manipulation.

\begin{figure*}[htbp]
    \centering
    \includegraphics[width=1\linewidth]{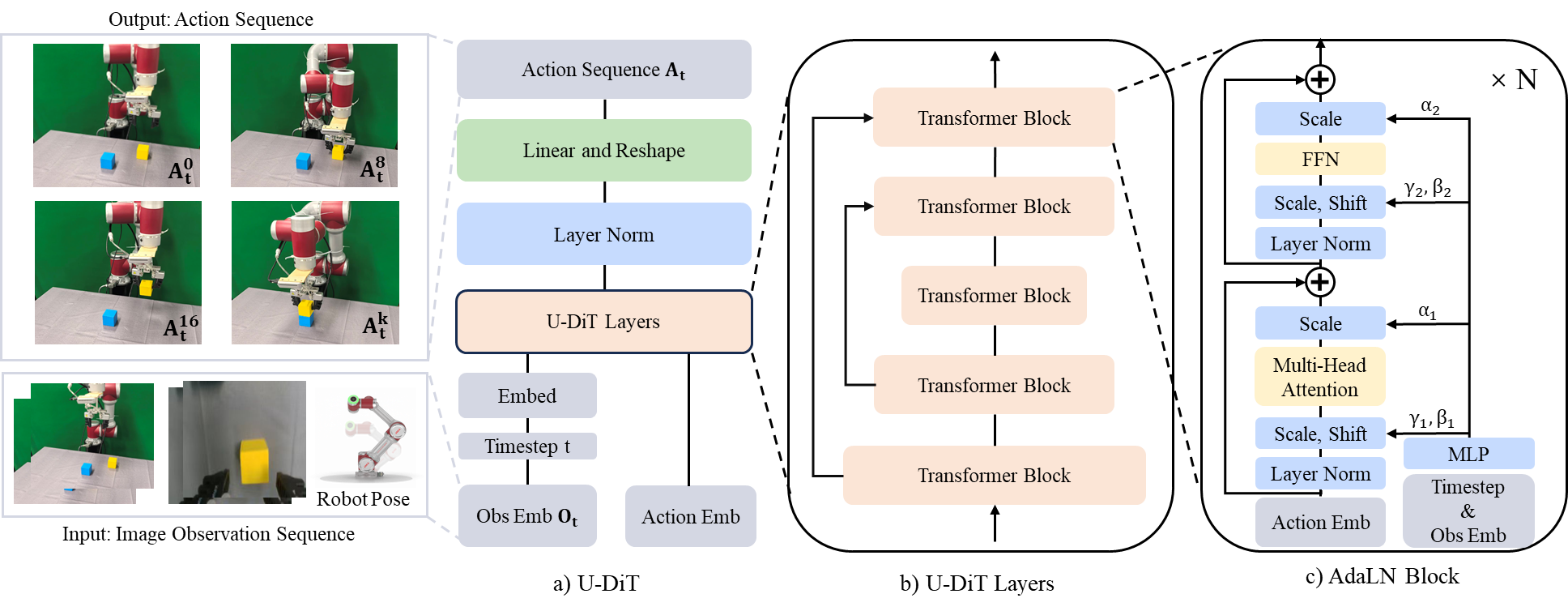}
    \caption{\textbf{Overview of U-DiT policy.} a) The overall structure of U-DiT is similar to DiT, but the connections between multiple DiT blocks are replaced with U-DiT layers. The policy takes the most recent $T_o$ steps of observation data $O_t$ as input and outputs the action sequence $A_t$. b) The specific form of the U-DiT layers. The width of each Transformer Block indicates its embedding dimension. c) The AdaLN block employs adaptive layer normalization to fuse conditions into the noisy action embeddings, resulting in more stable training and improved inference performance.
}
    \label{main}
\end{figure*}

\end{abstract}

\section{Introduction}
Imitation learning \cite{argall2009survey} has emerged as a prominent data-driven and sample-efficient approach for programming robots from expert demonstrations. Within this paradigm, behavior cloning is typically formulated as a supervised regression task that maps observations to corresponding actions. Previous studies explored various strategies for learning behavior cloning policies, such as directly predicting actions with regression models \cite{jang2022bc} or employing implicit policies \cite{florence2022implicit}. More recently, denoising diffusion probabilistic models (DDPMs) \cite{ho2020denoising} have achieved remarkable progress across multiple domains, including images, audio, video, and 3D generation \cite{song2020denoising, saharia2022photorealistic, richter2023speech, brooks2024video, lee2024dreamflow}. Their strength lies in capturing complex, multimodal distributions, which makes them particularly well-suited for robotics tasks where action spaces are inherently stochastic and multimodal. In this regard, diffusion-based policies have shown clear advantages over traditional regression-based cloning, as they alleviate issues of mode collapse and improve generalization. A representative example is Diffusion Policy \cite{chi2023diffusion}, which conditions the diffusion process on multimodal states, including visual observations, and has demonstrated strong empirical results. Notably, its variant Diffusion Policy in U-Net architecture (DP-U) has gained popularity in practice, as it performs well on diverse off-the-shelf tasks with minimal hyperparameter tuning. 

Despite these successes, DP-U still struggles to integrate global contextual information, which limits its ability to capture long-range dependencies in action sequences. Moreover, temporal convolutions inherently exhibit an inductive bias toward low-frequency signals \cite{tancik2020fourier}, which hinders performance on rapidly changing actions and causes over-smoothing. To mitigate this issue, Diffusion Policy has introduced a Transformer-based DDPM (DP-T), which alleviates over-smoothing by leveraging the global modeling capacity of Transformers. Methods such as \cite{zhu2025scaling, dasari2025ingredients} further enhance performance and scalability by replacing the cross-attention block \cite{vaswani2017attention} in DP-T with an AdaLN block \cite{peebles2023scalable}. These advances integrate the stepwise generative strengths of diffusion models with the global dependency modeling of Transformers, thereby improving the capture of long-range dependencies and yielding higher-quality trajectories.

However, while DP-T and its extensions benefit from global context modeling, they give up the U-shaped structure of U-Net \cite{ronneberger2015u}, which is known for its strong multi-scale feature fusion capability. To address the limitation of DP-U in global information modeling while retaining the multi-scale advantages of the U-shaped design, we propose U-shaped Diffusion Transformer (U-DiT) Policy. Inspired by the success of U-shaped Diffusion Transformers in image generation \cite{tian2024u}, U-DiT embeds Transformers into every layer of the U-Net, employing attention mechanisms to strengthen global contextual modeling. At the same time, the U-shaped structure preserves hierarchical multi-scale representations. Consequently, U-DiT achieves superior performance on long-horizon tasks, delivering improved accuracy and robustness in complex real-world scenarios and expanding the scope of potential applications. We validate our approach on 12 RLBench simulated tasks \cite{james2020rlbench} and further evaluate it on 4 real-world robotic tasks. Results from both simulation and real-robot experiments demonstrate that the U-DiT Policy consistently outperforms baseline diffusion strategies as well as DiT-based diffusion policies of comparable model size. The main contributions of this study are summarized as follows.

(1) We propose a novel U-DiT backbone. We incorporate U-shaped Diffusion Transformers into the Diffusion Policy framework and design an asymmetric decoder. By increasing the embedding dimension in the final layer, the model enhances its action representation capacity, resulting in improved performance on fine-grained robotic manipulation tasks.  

(2) We present an enhanced conditional modeling strategy. We integrate Adaptive Layer Normalization (AdaLN) and bidirectional attention masks to enable efficient fusion of observation conditions and temporal embeddings. This design captures global temporal dependencies during action diffusion, leading to higher training stability and more consistent action generation.  

(3) We perform extensive experiments on large-scale simulation and real-world robotic platforms, including visual generalization tests under varying lighting conditions and with distractor settings. Results show that our method significantly outperforms DP-U in terms of robustness and generalization, demonstrating strong practicality and application potential. 

\section{Related Work}
\subsection{Diffusion Models}
Diffusion models \cite{ho2020denoising,song2020denoising,song2020improved} have emerged as a powerful class of generative models, offering stable training by learning the score function and generating samples via iterative denoising. Their ability to model complex, multimodal distributions makes them well-suited for visuomotor policy learning.

The core idea of diffusion models is to generate samples by progressively mapping Gaussian noise to a target distribution. During training, the original data are gradually corrupted with noise in the forward process, while the model learns the reverse process to recover the data distribution from noise. Through iterative denoising, the model ultimately generates samples that approximate the original distribution. The standard Denoising Diffusion Probabilistic Model (DDPM) \cite{ho2020denoising} models the conditional distribution with a Gaussian: \begin{equation} p_\theta(x_{k-1} \mid x_k) = \mathcal{N}(x_{k-1} \mid \mu_\theta(x_k, k), \Sigma_k), \end{equation} where $\mu_\theta$ is predicted by a neural network. The training objective is to minimize the mean squared error (MSE) between the predicted and true noise.

\subsection{Diffusion Models in Robotics}
Recent works have adapted diffusion models to robotic control. Diffusion Policy \cite{chi2023diffusion} showed that diffusion models can be applied to learn state- or observation-conditioned action sequences, achieving robust imitation learning across diverse manipulation tasks. Building on this foundation, ScaleDP \cite{zhu2025scaling} scaled diffusion transformers to the billion-parameter regime, demonstrating that larger model capacity and efficient conditional representations substantially improve visuomotor control. $\pi$0 \cite{black2024pi} introduced a generalist policy architecture that leverages vision–language model (VLM) pre-training with flow matching, enabling robots to execute complex, language-conditioned tasks. DiT-Block Policy \cite{dasari2025ingredients} further improved long-horizon manipulation by proposing scalable attention blocks with adaptive layer normalization (AdaLN) and carefully initialized parameters, which stabilized the training of large diffusion transformers.

Despite these advances, existing diffusion-based policies still struggle with global information modeling and multi-scale feature fusion, limiting their effectiveness on long-horizon and complex real-world tasks. In contrast, we propose a novel diffusion backbone with a U-shaped architecture, which more effectively captures global context and hierarchical multi-scale representations, thereby enhancing both robustness and accuracy in visuomotor policy learning.

\subsection{Problem Setup}
We assume the existence of a demonstration dataset collected by an expert, denoted as $\mathcal{D} = \{\tau_0, \tau_1, \dots, \tau_n\}$. Each trajectory $\tau_i = \{(o_j, x_j)\}$ consists of a sequence of visual observations $o_j$ and proprioceptive states $x_j$. The proprioceptive information can include the pose or joint angles of the manipulator, as well as the open/close state of the gripper. In this work, we use the 6D end-effector pose, that is, position $(x, y, z)$ and rotation $(roll, pitch, yaw)$, as the control signal.

\section{METHOD}
\subsection{Diffusion Policy}
Diffusion Policy (DP) \cite{chi2023diffusion} formulates robotic action generation as a conditional denoising process, enabling the capture of multimodal action distributions. It models the conditional distribution $p(A_t \mid o_t, x_t)$, where $A_t = \{a_t, \dots, a_{t+C}\}$ is the future $C$-step action sequence conditioned on observation $o_t$ and state $x_t$. During sampling, $A_t^K$ is initialized with Gaussian noise and iteratively denoised to yield $A_t^0$:
\begin{equation}
A_t^{k-1} = \alpha_k A_t^k - \gamma_k \, \epsilon_\theta(o_t, x_t, A_t^k, k) + \mathcal{N}(0, \sigma_k^2 I),
\end{equation}
where $\epsilon_\theta$ is the noise prediction network and $(\alpha_k, \gamma_k, \sigma_k)$ are diffusion parameters. The training objective minimizes the MSE between predicted and true noise:
\begin{equation}
L(\theta) = \mathbb{E}\left[\| \epsilon_k - \epsilon_\theta(o_t, x_t, A_t^k + \epsilon_k, k) \|^2 \right].
\end{equation}

DP consists of a state encoder, an action encoder, and an action decoder. The state encoder extracts conditional features $h_t$, the action encoder models the noisy sequence into $X_t^k$, and the decoder combines them to predict residual noise using Feature-wise Linear Modulation (FiLM) \cite{perez2018film}. At inference, trajectories are generated by iterative denoising and executed in a receding-horizon closed-loop manner.

\begin{figure*}
    \centering
    \includegraphics[width=1\linewidth]{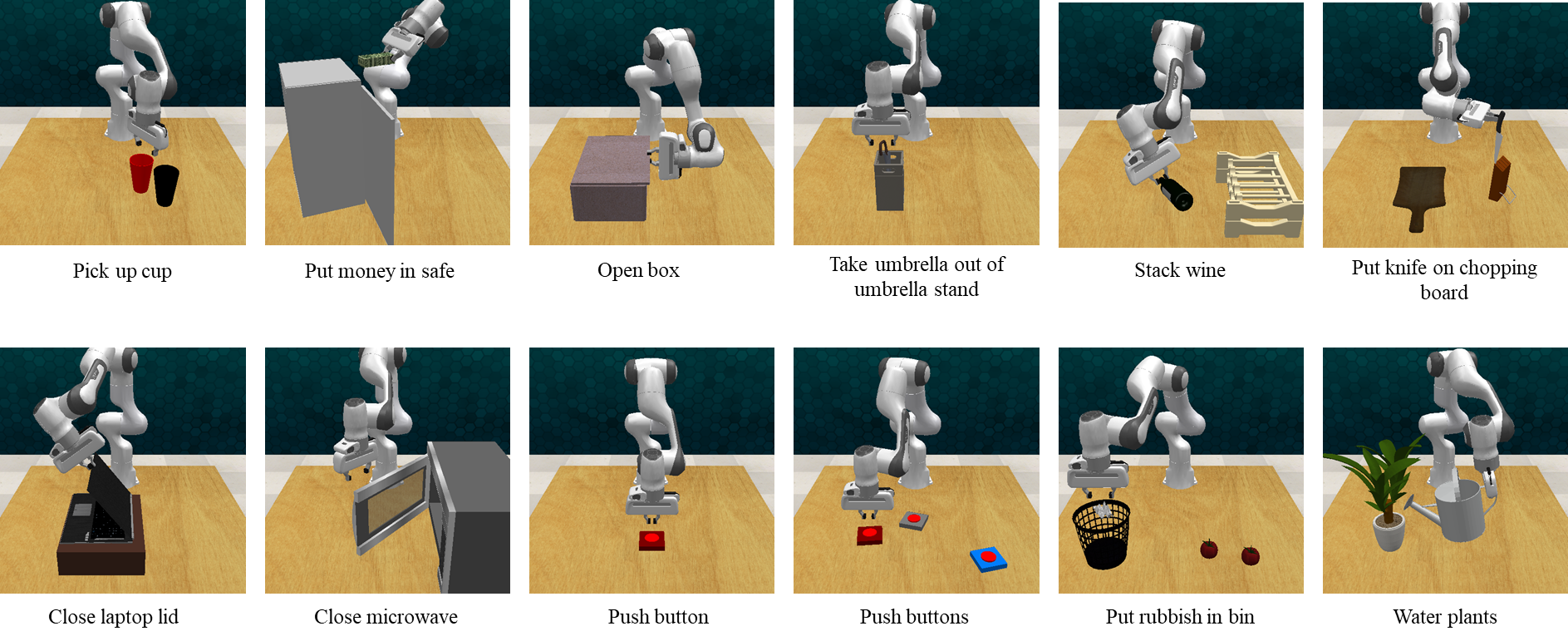}
    \caption{\textbf{RLBench tasks.} We conduct experiments on 12 simulated tasks in RLBench.}
    \label{RLBench task}
\end{figure*}

\subsection{U-DiT Policy}
In this study, we adopt the U-shaped Diffusion Transformers (U-DiT) as the backbone network for action diffusion policies. U-DiT takes inspiration from the U-Net architecture in image generation. Through a symmetric encoder–decoder design, it introduces multi-scale feature fusion and performs conditional denoising prediction on noisy action sequences at each step of the diffusion process. Unlike a standard Transformer, U-DiT applies progressive downsampling and upsampling along the temporal dimension to establish global dependencies. Skip connections are used to preserve local details, which match the requirements of robotic manipulation tasks where action prediction needs both long-term consistency and fine-grained control.

Specifically, we embed DiT blocks into a canonical U-Net design. Following the standard U-Net structure, U-DiT consists of an encoder and a decoder with the same number of levels. The encoder downsamples the input action sequence into stage-level representations, while the decoder progressively upsamples the encoded sequence back to the original resolution.

At each encoder stage transition, spatial downsampling is performed with a factor of 2, and the feature dimension is doubled. As shown in Fig. \ref{main}, each DiT layer in a stage is made up of multiple DiT blocks, whose sizes change with down-sampling or up-sampling across stages. Skip connections are provided at every stage transition. The skipped features are concatenated and fused with the upsampled output of the previous decoder stage. This design allows action representations to unfold across multiple scales and compensates for the information loss caused by downsampling in the encoder.

Given an action space (32×7), we specify a total of three stages. Features are downsampled twice and then restored to their original size. To align temporal and conditional embeddings with different feature dimensions at multiple scales, we use independent embedding modules for each stage. In addition, we avoid patching the latent space, since the U-Net architecture already downsamples it and no further spatial compression is needed. In addition, we adopt an asymmetric network design, where the embedding dimension of the final layer does not decrease but remains the same as that of the penultimate layer.
After processing with the final DiT block, the decoder output is mapped through a linear layer to obtain the noise prediction $\epsilon_\theta(O_t, A_t^k, k)$. This prediction is used in the denoising update of the diffusion reverse process, thus gradually generating the noise-free action sequence $A_t^0$.

The following provides a detailed description of each component in the U-DiT architecture.

\textbf{DiT Block.} Within each Transformer block of U-DiT, the noisy embeddings of the action sequence are first processed by Multi-Head Self-Attention to model temporal dependencies. To incorporate conditional information, we adopt Adaptive Layer Normalization (AdaLN) \cite{peebles2023scalable}. In this mechanism, the timestep embedding of diffusion step $k$ is combined with the observation features $O_t$ to produce a set of scaling factors $\gamma(k,O_t)$ and shifting factors $\beta(k,O_t)$, which are used to adjust the normalized output:
\begin{equation}
\text{AdaLN}(x) = (1 + \gamma(k, O_t)) \cdot \text{LN}(x) + \beta(k, O_t),
\end{equation}
where $\text{LN}(\cdot)$ denotes standard layer normalization. Unlike pure cross-attention, AdaLN allows conditional information to directly modulate the distribution of action representations, which helps stabilize gradients during training and improves prediction performance.

\textbf{Asymmetric  Design.} In the decoder part, we introduce further improvements. Leveraging the advantages of a decoder-only architecture, we shift the computational focus from the encoder stage (the downsampling half of the model) to the decoder stage (the upsampling half of the model). Specifically, the embedding dimension of the final layer no longer decreases but remains the same as that of the penultimate layer. This modification enhances the representational capacity on the output side, enabling the model to better preserve fine-grained features when generating action sequences and preventing the loss of control performance caused by excessive compression. This design is consistent with previous studies \cite{vaswani2017attention, hoogeboom2023simple}, which have shown that decoder-dominant architectures often achieve superior performance in generative tasks.

\textbf{Bidirectional Attention Mask.} Following the Transformer architecture proposed by \cite{vaswani2017attention}, Diffusion Policy adopts causal attention mask, which restricts each position to attend only to historical information. However, in this work, we do not generate actions step by step. Instead, we perform holistic prediction on a compact trajectory representation ($32 \times 7$). For such sequence-level modeling, a causal attention mask limits information exchange across different timesteps. To address this issue, we employ a bidirectional attention mask that allows any timestep in the sequence to attend to all others, thereby fully leveraging global context during the diffusion process. This design is inspired by ScaleDP \cite{zhu2025scaling} and $\pi$0 \cite{black2024pi}, both of which have demonstrated the advantages of bidirectional masking in capturing complex dependencies for large-scale diffusion policies and language–action modeling.

\begin{figure*}[htbp]
    \centering
    \includegraphics[width=1\linewidth]{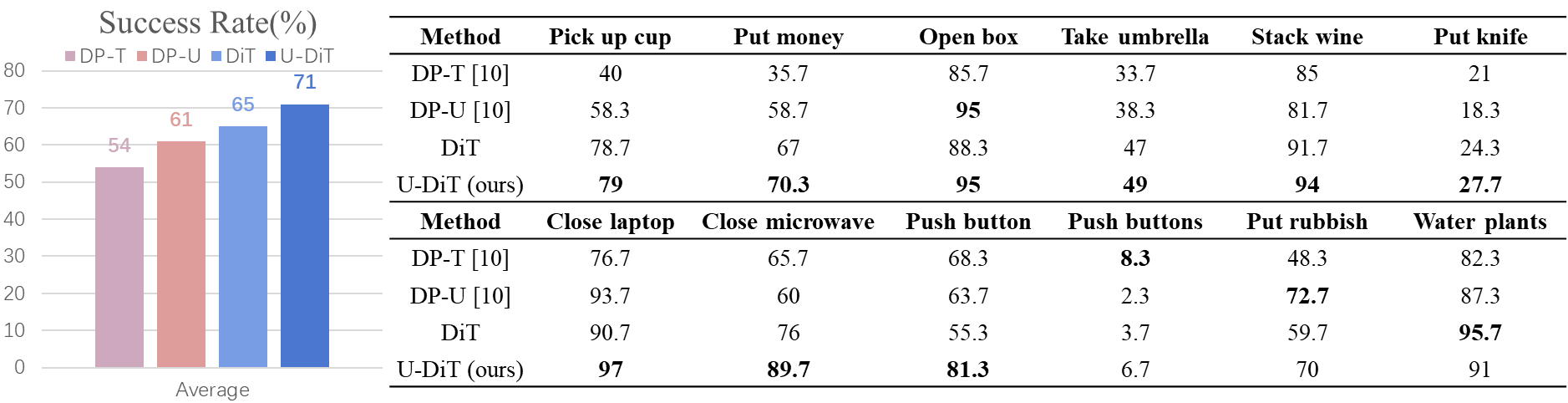}
    \caption{\textbf{Sim evaluation.} We compare U-DiT Policy with the original DP and DP with a DiT backbone across 12 RLBench tasks. U-DiT outperforms both DiT and DP on most tasks, achieving an average success rate that is 10\% higher than DP-U and 6\% higher than DiT.}

    \label{RLBench result}
\end{figure*}

\section{EXPERIMENTS}
In our experiments, we demonstrate the effectiveness of U-DiT Policy from two perspectives: (1) performance and data scalability compared with existing diffusion policies, and (2) robustness to variations in visual observations, including changes in appearance, objects, lighting, and distractors.  

\subsection{Simulation Experiments}
\textbf{RLBench.} RLBench \cite{james2020rlbench} is a benchmark for robotic tasks. All tasks defined in RLBench involve controlling a simulated 7-DOF Franka Emika Panda robotic arm with a parallel gripper to perform diverse object rearrangement tasks under randomized initial conditions. Each task in the RLBench suite provides its own binary success criterion, most of which determine whether a specific object or set of objects has been rearranged within a positional threshold of the target location.  

\textbf{Experimental Setup.} In this study, we adopt a dual-view setting on RLBench using both a front camera and a wrist camera. We evaluate 12 tasks, with visualizations shown in Fig. \ref{RLBench task}. Each method is trained with 50 demonstrations, and every checkpoint is evaluated on 100 unseen episodes per task with three random seeds. Due to its training difficulty, DP-T is trained for 3000 epochs, while all other models are trained for 1500 epochs to ensure convergence. Checkpoints are saved every 50 epochs, and the top three checkpoints with the highest success rates are selected to compute the average. All models in both simulation and real-world experiments are trained on NVIDIA A100 GPUs. For inference and evaluation, we use an NVIDIA GeForce RTX 4060 Ti.

\textbf{Comparison with Diffusion Policies.} We compare U-DiT Policy with DP-T, DP-U, and DiT-based backbones, where DiT and U-DiT have a comparable number of parameters (93M). Detailed descriptions of these baselines are provided in the real-world experiment section. As shown in Fig. \ref{RLBench result}, our model achieves higher success rates in most tasks. In particular, on the Close Microwave and Push Button tasks, U-DiT shows the largest margins, surpassing DP-T by 24\% and 13.2\%, DP-U by 29.7\% and 17.6\%, and DiT by 13.7\% and 26\%, respectively. Averaged over all 12 tasks, U-DiT achieves performance improvements of 17\% over DP-T, 10\% over DP-U, and 6\% over DiT.

\textbf{Data Scaling.} Inspired by previous studies on data scaling \cite{kaplan2020scaling, yu2023scaling}, we investigate how U-DiT scales when trained with increasingly limited demonstration data. Since U-DiT is structurally similar to DiT, we focus on highlighting the unique benefits of the U-shaped design by training both models with only 10 demonstrations on RLBench. As reported in Table~\ref{tab:rlbench}, U-DiT outperforms DiT on most tasks under this low-data regime. On average, U-DiT maintains a success rate of 23\%, compared to 18\% for DiT, resulting in a 5\% margin.

\begin{table}[htbp]
\centering
\caption{Data Scaling Results on RLBench}
\label{tab:rlbench}
\resizebox{\columnwidth}{!}{%
\begin{tabular}{lcccc}
\toprule
\textbf{Method} & \textbf{Pick up cup} & \textbf{Put money} & \textbf{Open box} & \textbf{Take umbrella} \\
\midrule
DiT    & 12.7 & 23.3   & 21  & 5.7 \\
U-DiT (ours)  & \textbf{19.7}   & \textbf{31.7} & \textbf{29.7}   & \textbf{8.7} \\
\midrule
\textbf{Method} & \textbf{Stack wine} & \textbf{Put knife} & \textbf{Close laptop lid} & \textbf{Close microwave} \\
\midrule
DiT    & 25.3 & 6 & \textbf{45.3} & 42.7 \\
U-DiT (ours)  & \textbf{26}   & \textbf{7.7} & \textbf{45.3}   & \textbf{51} \\
\midrule
\textbf{Method} & \textbf{Push button} & \textbf{Push buttons} & \textbf{Put rubbish} & \textbf{Water plants} \\
\midrule
DiT    & 19.3 & 3.7 & 3.3 & 11.3 \\
U-DiT (ours)  & \textbf{19.3} & \textbf{6.7} & \textbf{11} & \textbf{20} \\
\bottomrule
\end{tabular}%
}
\end{table}
\subsection{Real Robot Experiments}
\begin{figure}
    \centering
    \includegraphics[width=1\linewidth]{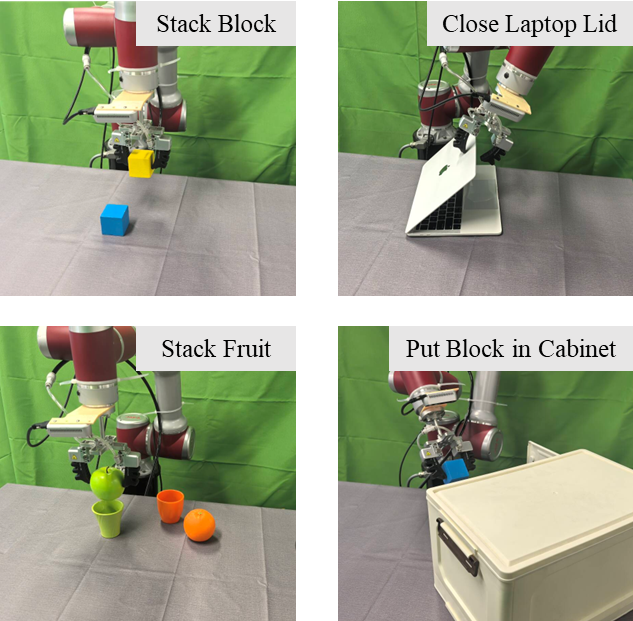}
    \caption{\textbf{Real-world tasks on Jaka robot}. Task 1: \textit{Stack Block}, Task 2: \textit{Close Laptop Lid}, Task 3: \textit{Stack Fruit}, Task 4: \textit{Put Block in Cabinet}. Each task is evaluated with 20 trials. }
    \label{fig:real}
\end{figure}

\textbf{Real-world Setup.}
Our U-DiT Policy is evaluated on four tasks using a 6-DOF Jaka robotic arm equipped with a pneumatic gripper. We use two RealSense D435 cameras to capture visual observations in the real world. As illustrated in Fig. \ref{fig:real}, a brief description of our tasks is as follows.

\textbf{Data Collection.} We collect datasets through human demonstrations, where the robot arm is teleoperated using a SpaceMouse. For each task, objects are randomly placed within a designated area, and the human operator is instructed to manipulate them as smoothly as possible. For the \textit{put block in cabinet} task, the cabinet is oriented toward the right or rear side to prevent interference from data cables. During demonstrations, we record RGB streams from two different viewpoints: a wrist-mounted camera and a front-facing camera. In addition, we capture the robot states, including joint positions and end-effector poses. U-DiT follows a common control mode and predicts 6D actions, consisting of position $(x, y, z)$ and rotation $(roll, pitch, yaw)$. For each task, we collect 50 trajectories. For the task of \textit{Close Laptop Lid}, we collect 30 trajectories.

\textbf{Baselines.} Firstly, we compare U-DiT with SOTA baselines to evaluate its performance. These baselines include:  
\textbf{(1) Action Chunking Transformers (ACT) \cite{zhao2023learning}:} 
ACT adopts a standard encoder–decoder Transformer architecture, specifically DeTR \cite{carion2020end}. The encoder processes input observation tokens, which include camera observations (encoded with ResNet18), a goal-conditioning vector, and (optionally) a latent plan vector computed from ground-truth actions during training (randomly sampled during inference). The network is optimized with behavior cloning (BC) using an $L_1$ regression loss on expert actions. We implement this baseline with the recommended hyperparameters and omit the latent plan vector, following the authors’ guidelines.  \textbf{(2) DP-U and DP-T \cite{chi2023diffusion}}:  DP-U is the original implementation of Diffusion Policy, where camera observations are processed into feature maps using separate ResNets, reduced to vectors through spatial softmax \cite{levine2016end}, and then fed into a conditional U-Net serving as the noise prediction network. DP-T follows a similar setup, but replaces the U-Net noise network with a Transformer using standard causal cross-attention blocks. Specifically, for DP-U we follow its default parameter size of 256M in RoboMimic \cite{mandlekar2022matters} for simulation experiments. In real-world experiments, its default size is 64M, which we scale up to 93M to match the parameter size of our model, enabling a fair comparison. \textbf{(3) Diffusion Transformer (DiT).}  Since the architecture of U-DiT is similar to DiT, we include DiT-based diffusion policy as a baseline for direct comparison. Although DiT adopts a pure Transformer design to model long-range dependencies, it lacks the U-shaped multi-scale feature fusion that has proven highly effective in image generation tasks. By introducing U-shaped layers, U-DiT combines the global modeling ability of Transformers with the multi-scale representation power of U-Net. Specifically, we set the DiT parameters following the ScaleDP configuration \cite{zhu2025scaling}, scaling the model size to 93M to match our model for a fair comparison.

\textbf{Task Setup.} Our real-world experiments include four tasks, consisting of two relatively simple single-stage tasks and two complex multi-stage long-horizon tasks. Below we describe each task and its success criterion in detail. \textbf{(1) Stack Block:} A blue block and a yellow block are randomly placed within the robot’s workspace. The robot must first grasp the yellow block and place it on top of the blue block. The task is considered successful if the yellow block remains on the blue block after the gripper is released.  \textbf{(2) Close Laptop Lid:} A laptop is randomly placed in the designated area with its lid open. The robot must grasp the lid and close it. The task is considered successful if, after the gripper is released, the angle between the lid and the keyboard is less than $30^\circ$.  \textbf{(3) Stack Fruit:} A green apple, an orange, a green cup, and an orange cup are randomly placed within the designated area. This is a two-stage task: the robot must first place the green apple into the green cup, and then place the orange into the orange cup. The task is considered successful if both placements are completed in order.  \textbf{(4) Put Block in Cabinet:} This is a multi-stage task. A blue block and a cabinet are randomly placed in the designated area. The robot must first open the cabinet door and then place the blue block inside. The task is considered successful if the block is deposited within the cabinet.  

\textbf{Experimental Results.} Table \ref{table2} presents the real robot experimental results. It can be seen that U-DiT consistently outperforms DP-U in all tasks, with an average improvement of 22.5\%. Compared with DiT, U-DiT achieves a 17.5\% gain, with the margin further enlarged relative to simulation tasks, highlighting the advantage of the U-shaped design. Moreover, compared to ACT, a strong baseline for imitation learning , U-DiT delivers an average improvement of 18.75\% in four tasks. These results demonstrate the effectiveness and superiority of U-DiT in both robustness and performance.

\begin{table}[ht]
\caption{Results of Comparative Studies}
\begin{center}
\begin{tabular}{lccccc}
\toprule
\textbf{Model} & \textbf{Task1} & \textbf{Task2} & \textbf{Task3} & \textbf{Task4} & \textbf{Avg.} \\
\midrule
ACT \cite{zhao2023learning} & 80 & 65 & 35 & 40 & 55 \\
DP-T \cite{chi2023diffusion} & 70 & 60 & 30 & 35 & 48.75 \\
DP-U \cite{chi2023diffusion} & 75 & 60 & 30 & 40 & 51.25 \\
DiT & 80 & 65 & 35 & 45 & 56.25 \\
U-DiT (ours) & \textbf{95} & \textbf{85} & \textbf{50} & \textbf{65} & \textbf{73.75} \\
\bottomrule
\end{tabular}
\label{table2}
\end{center}
\end{table}

\textbf{Ablations.} We compare the DiT-block policy against three ablations that use the same exact setup but replace specific modules: \textbf{(1) Cross-Attention:} The diffusion decoder uses standard per-layer cross-attention blocks \cite{vaswani2017attention} to modulate the memory embeddings from the encoder stack. \textbf{(2) Symmetric Structure:}  
As mentioned in the previous section, U-DiT keeps the embedding dimension of the final layer equal to that of the penultimate layer. In this ablation, we instead use a symmetric structure in which the embedding dimension of the final layer is half of the penultimate layer. \textbf{(3) Causal Attention:}  Causal attention ensures that the model can only attend to past actions and observations, preventing it from accessing information from the future.  

As given in Table \ref{ablations}, our model with an asymmetric structure achieves an improvement of 6.25\% over the symmetric design. In contrast, replacing AdaLN with cross-attention blocks leads to larger performance degradation, consistent with previous findings in DP \cite{chi2023diffusion} and ScaleDP \cite{zhu2025scaling} that DP-T is highly sensitive to parameterization and difficult to scale. Moreover, the model with causal attention performs comparably on short-horizon tasks but suffers significant drops on multistage tasks, especially in the middle and later phases, where abnormal oscillations often occur. This observation further supports that, for action sequence modeling, bidirectional attention is more effective in mitigating temporal inconsistencies across successive actions.
\begin{table}
\caption{Ablations}
\begin{center}
\begin{tabular}{lccccc}
\toprule
\textbf{Model} & \textbf{Task1} & \textbf{Task2} & \textbf{Task3} & \textbf{Task4} & \textbf{Avg.} \\
\midrule
w/o asymmetric  & 90 & 75 & \textbf{50} & 60 & 67.5 \\
cross-attention block & 65 & 60 & 30 & 35 & 47.5 \\
causal attention  & \textbf{95} & \textbf{85} & 30 & 50 & 65 \\
U-DiT (ours) & \textbf{95} & \textbf{85} & \textbf{50} & \textbf{65} & \textbf{73.75} \\
\bottomrule
\end{tabular}
\label{ablations}
\end{center}
\end{table}
\begin{figure}
   \centering
   \includegraphics[width=1\linewidth]{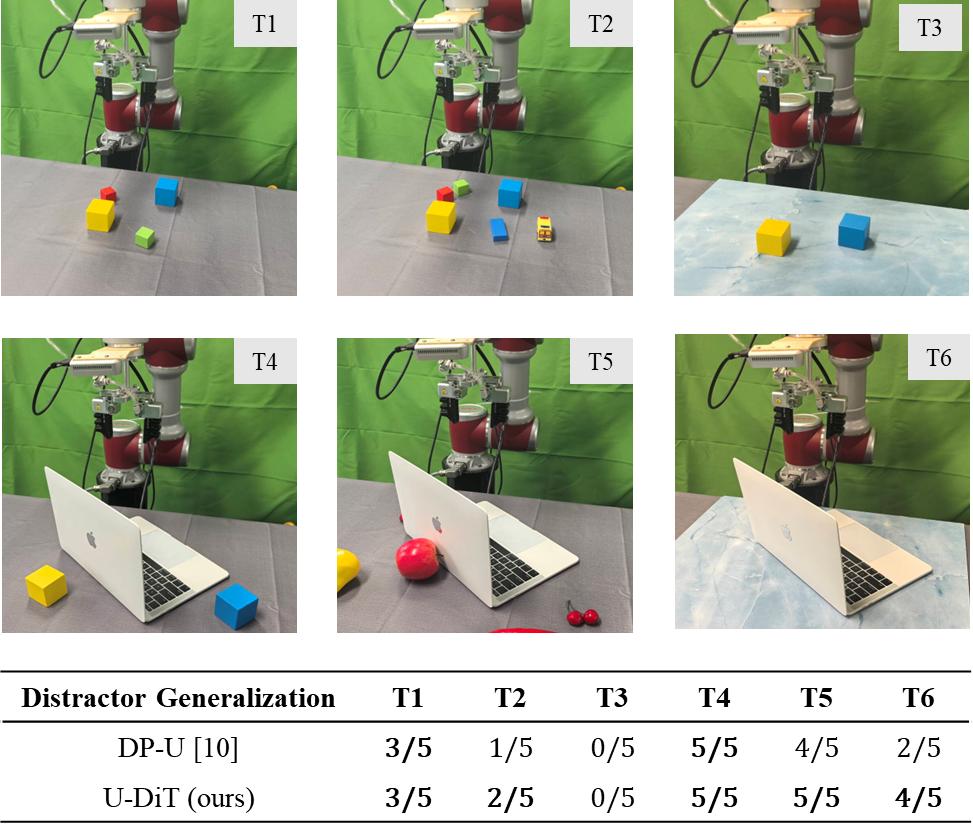}
   \caption{\textbf{Distractor generalization}. We test these capabilities on \textit{Stack Block} and \textit{Close Laptop Lid} tasks. }
   \label{fig:distractor}
\end{figure}

\textbf{Distractor Generalization.} To evaluate the robustness of the policy in complex scenarios, we introduce additional distractor objects during the testing phase. Specifically, we add objects with diverse shapes and colors (e.g., fruits, extra blocks, cluttered table backgrounds) around the manipulation area to simulate common disturbances in real-world environments. Example experimental scenes are shown in Fig.~\ref{fig:distractor} (T1–T6), where T1–T3 correspond to the \textit{stack block} task and T4–T6 correspond to the \textit{close laptop lid} task. To ensure reliability and avoid damage caused by accidental jitter, each distractor scenario is evaluated five times, with different object placements under the same distractor setting. We also confirm that under no-distractor conditions, both methods are able to complete the tasks successfully. In T1, T4, and T5, which are relatively less distracting scenarios, both methods achieve high success rates and perform comparably. However, as the level of distraction increases, for example, in T2,  where distractors share similar colors, DP-U tends to deviate when grasping the yellow block. In contrast, our U-DiT, though only slightly better overall, consistently aligns with the yellow block in the first grasping stage, with failures occurring mainly during placement onto the blue block. In T3, where the table background is changed, neither method succeeds; however, U-DiT still manages to align with the yellow block during grasping, whereas DP-U completely misses it. In T6, although the background is also altered, the large footprint of the laptop reduces the impact of background distractors. In this case, U-DiT shows better height control when grasping and slowly closing the laptop lid, while DP-U frequently applies excessive downward force, leading to damage of the laptop.

\begin{figure}
    \centering
    \includegraphics[width=1\linewidth]{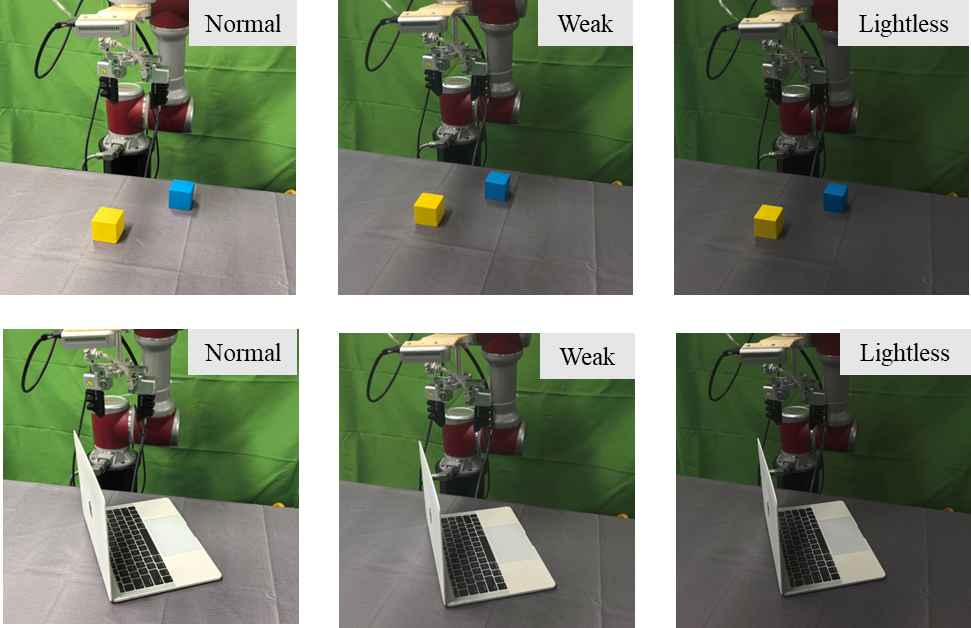}
    \caption{\textbf{Light generalization.} We evaluate on the \textit{Stack Block} and \textit{Close Laptop Lid} tasks.}
    \label{light1}
\end{figure}

\textbf{Light Generalization.} Changes in lighting conditions are common challenges for robots operating in real-world environments. As depicted in Fig.~\ref{light1}, we present three lighting levels, i.e., normal illumination (Normal), weak illumination (Weak), and complete darkness (Lightless). By varying the lighting intensity, we evaluate the stability of the model under changes in image brightness and contrast. To ensure reliability and avoid damage caused by accidental jitter, each lighting condition is evaluated across five trials, with different object placements under the same lighting setting. We also verify that under normal illumination without distractors, both methods are able to complete the tasks successfully. The results are given Table~\ref{tab:light_generalization}. In the \textit{stack block} task, U-DiT was able to complete the task 4 times under weak illumination, whereas DP-U only succeeded twice. Since stack block requires high precision, neither model could accomplish the task under lightless conditions. However, the U-DiT gripper approached the yellow block with greater precision, while DP-U showed a significant deviation. In the \textit{close laptop lid} task, U-DiT completed the task 3 times under weak illumination, compared to only 1 success by DP-U. In addition, DP-U often pressed the gripper excessively, causing damage to the prop. Under lightless conditions, only U-DiT managed to barely complete the task once.

\begin{table}[ht]
\caption{Light Generalization Results}
\begin{center}
\begin{tabular}{lccc}
\toprule
\textbf{Stack Block} & Normal & Weak & Lightless \\
\midrule
DP-U \cite{chi2023diffusion}   & \textbf{5/5} & 2/5 & 0/5 \\
U-DiT (ours) &\textbf{5/5}& \textbf{4/5} & 0/5 \\
\midrule
\textit{Close Laptop Lid} & Normal & Weak & Lightless \\
\midrule
DP-U \cite{chi2023diffusion}   & \textbf{5/5} & 1/5 & 0/5 \\
U-DiT (ours)  & \textbf{5/5} & \textbf{3/5} & \textbf{1/5} \\
\bottomrule
\end{tabular}
\label{tab:light_generalization}
\end{center}
\end{table}
\section{Discussion}
Compared with DP-U, U-DiT shows advantages in capturing long-horizon dependencies and maintaining more coherent multi-stage action sequences. While DP-U leverages convolutional encoders and decoders to fuse multi-scale features, its limited receptive field makes it less effective at modeling global task context, often leading to incoherent action sequences in multi-stage tasks. U-DiT addresses this shortcoming by integrating self-attention into every stage of the U-shaped architecture, thereby combining the fine-grained precision of DP-U with the global consistency of transformer-based policies. Moreover, U-DiT exhibits stronger robustness to visual distractors, as the joint use of multi-scale representations and global attention allows the policy to suppress irrelevant scene variations while preserving task-relevant details. This dual strength contributes to the improved accuracy and robustness observed in both simulation and real-world benchmarks.
 
Compared with DP-T and its extensions, such as the DiT-based diffusion policy, our U-DiT tends to achieve stronger performance on long-horizon manipulation tasks under similar parameter counts. While DP-T benefits from pure Transformer layers that enable powerful global context modeling, it discards the U-shaped architecture of DP-U, thereby losing explicit multi-scale feature fusion. Experimental results indicate that this trade-off may limit performance in settings where both global context and fine-grained details are important.

By incorporating the U-shaped design, U-DiT retains hierarchical feature representations through encoder–decoder pathways, which are essential for preserving local action details such as precise gripper trajectories. At the same time, transformer-based attention layers embedded throughout the U-Net enhance global reasoning, helping to improve stage-to-stage consistency in multi-step tasks. This combination allows U-DiT to achieve somewhat higher overall success rates than the DiT-based diffusion policy in our experiments. The data scaling results further suggest that U-DiT tends to perform better with fewer training trajectories, indicating a potential advantage in leveraging limited demonstrations.

Despite the demonstrated improvements, this study has certain limitations. First, the evaluation is restricted to table-top manipulation tasks. It remains unclear how well the approach generalizes to more dynamic or unstructured environments. Second, the iterative denoising process introduces inference latency, which poses challenges for real-time deployment.

\section{Conclusion}
In this study, we proposed a novel diffusion policy network based on the U-shaped Diffusion Transformer (U-DiT) for action prediction in robotic manipulation tasks. Across a variety of robotic manipulation benchmarks, the proposed method achieves higher success rates than DP, with an average improvement of 22.5\% in real-world tasks. In addition, visual generalization experiments validate the robustness of the model in complex environments. The results show that U-DiT outperforms DP-U under varying lighting conditions and distractor interference, maintaining focus on target objects, and resisting perceptual disturbances within a practical range. These findings highlight the advantages of our method in enhancing generalization ability and practical applicability.

Future work will focus on two directions: (1) exploring more efficient sampling and inference methods to further reduce control latency; (2) incorporating multi-modal sensing (e.g., touch and force feedback) to improve performance in tasks involving complex contact dynamics.   

\bibliographystyle{IEEEtran}
\bibliography{reference}

\begin{thebibliography}{10}
\providecommand{\url}[1]{#1}
\csname url@samestyle\endcsname
\providecommand{\newblock}{\relax}
\providecommand{\bibinfo}[2]{#2}
\providecommand{\BIBentrySTDinterwordspacing}{\spaceskip=0pt\relax}
\providecommand{\BIBentryALTinterwordstretchfactor}{4}
\providecommand{\BIBentryALTinterwordspacing}{\spaceskip=\fontdimen2\font plus
\BIBentryALTinterwordstretchfactor\fontdimen3\font minus \fontdimen4\font\relax}
\providecommand{\BIBforeignlanguage}[2]{{%
\expandafter\ifx\csname l@#1\endcsname\relax
\typeout{** WARNING: IEEEtran.bst: No hyphenation pattern has been}%
\typeout{** loaded for the language `#1'. Using the pattern for}%
\typeout{** the default language instead.}%
\else
\language=\csname l@#1\endcsname
\fi
#2}}
\providecommand{\BIBdecl}{\relax}
\BIBdecl

\bibitem{argall2009survey}
B.~D. Argall, S.~Chernova, M.~Veloso, and B.~Browning, ``A survey of robot learning from demonstration,'' \emph{Robotics and Autonomous Systems}, vol.~57, no.~5, pp. 469--483, 2009.

\bibitem{jang2022bc}
E.~Jang, A.~Irpan, M.~Khansari, D.~Kappler, F.~Ebert, C.~Lynch, S.~Levine, and C.~Finn, ``{BC-Z: Zero-shot} task generalization with robotic imitation learning,'' in \emph{Conference on Robot Learning}.\hskip 1em plus 0.5em minus 0.4em\relax PMLR, 2022, pp. 991--1002.

\bibitem{florence2022implicit}
P.~Florence, C.~Lynch, A.~Zeng, O.~A. Ramirez, A.~Wahid, L.~Downs, A.~Wong, J.~Lee, I.~Mordatch, and J.~Tompson, ``Implicit behavioral cloning,'' in \emph{Conference on Robot Learning}.\hskip 1em plus 0.5em minus 0.4em\relax PMLR, 2022, pp. 158--168.

\bibitem{ho2020denoising}
J.~Ho, A.~Jain, and P.~Abbeel, ``Denoising diffusion probabilistic models,'' \emph{Advances in Neural Information Processing Systems}, vol.~33, pp. 6840--6851, 2020.

\bibitem{song2020denoising}
J.~Song, C.~Meng, and S.~Ermon, ``Denoising diffusion implicit models,'' \emph{arXiv preprint arXiv:2010.02502}, 2020.

\bibitem{saharia2022photorealistic}
C.~Saharia, W.~Chan, S.~Saxena, L.~Li, J.~Whang, E.~L. Denton, K.~Ghasemipour, R.~Gontijo~Lopes, B.~Karagol~Ayan, T.~Salimans \emph{et~al.}, ``Photorealistic text-to-image diffusion models with deep language understanding,'' \emph{Advances in Neural Information Processing Systems}, vol.~35, pp. 36\,479--36\,494, 2022.

\bibitem{richter2023speech}
J.~Richter, S.~Welker, J.-M. Lemercier, B.~Lay, and T.~Gerkmann, ``Speech enhancement and dereverberation with diffusion-based generative models,'' \emph{IEEE/ACM Transactions on Audio, Speech, and Language Processing}, vol.~31, pp. 2351--2364, 2023.

\bibitem{brooks2024video}
T.~Brooks, B.~Peebles, C.~Holmes, W.~DePue, Y.~Guo, L.~Jing, D.~Schnurr, J.~Taylor, T.~Luhman, E.~Luhman \emph{et~al.}, ``Video generation models as world simulators,'' \emph{OpenAI Blog}, vol.~1, no.~8, p.~1, 2024.

\bibitem{lee2024dreamflow}
K.~Lee, K.~Sohn, and J.~Shin, ``Dreamflow: High-quality text-to-3d generation by approximating probability flow,'' \emph{CoRR}, 2024.

\bibitem{chi2023diffusion}
C.~Chi, Z.~Xu, S.~Feng, E.~Cousineau, Y.~Du, B.~Burchfiel, R.~Tedrake, and S.~Song, ``Diffusion policy: Visuomotor policy learning via action diffusion,'' \emph{The International Journal of Robotics Research}, p. 02783649241273668, 2023.

\bibitem{tancik2020fourier}
M.~Tancik, P.~Srinivasan, B.~Mildenhall, S.~Fridovich-Keil, N.~Raghavan, U.~Singhal, R.~Ramamoorthi, J.~Barron, and R.~Ng, ``Fourier features let networks learn high frequency functions in low dimensional domains,'' \emph{Advances in Neural Information Processing Systems}, vol.~33, pp. 7537--7547, 2020.

\bibitem{zhu2025scaling}
M.~Zhu, Y.~Zhu, J.~Li, J.~Wen, Z.~Xu, N.~Liu, R.~Cheng, C.~Shen, Y.~Peng, F.~Feng \emph{et~al.}, ``Scaling diffusion policy in transformer to 1 billion parameters for robotic manipulation,'' in \emph{2025 IEEE International Conference on Robotics and Automation (ICRA)}.\hskip 1em plus 0.5em minus 0.4em\relax IEEE, 2025, pp. 10\,838--10\,845.

\bibitem{dasari2025ingredients}
S.~Dasari, O.~Mees, S.~Zhao, M.~K. Srirama, and S.~Levine, ``The ingredients for robotic diffusion transformers,'' in \emph{2025 IEEE International Conference on Robotics and Automation (ICRA)}.\hskip 1em plus 0.5em minus 0.4em\relax IEEE, 2025, pp. 15\,617--15\,625.

\bibitem{vaswani2017attention}
A.~Vaswani, N.~Shazeer, N.~Parmar, J.~Uszkoreit, L.~Jones, A.~N. Gomez, {\L}.~Kaiser, and I.~Polosukhin, ``Attention is all you need,'' \emph{Advances in Neural Information Processing Systems}, vol.~30, 2017.

\bibitem{peebles2023scalable}
W.~Peebles and S.~Xie, ``Scalable diffusion models with transformers,'' in \emph{Proceedings of the IEEE/CVF International Conference on Computer Vision}, 2023, pp. 4195--4205.

\bibitem{ronneberger2015u}
O.~Ronneberger, P.~Fischer, and T.~Brox, ``U-net: Convolutional networks for biomedical image segmentation,'' in \emph{International Conference on Medical Image Computing and Computer-Assisted Intervention}.\hskip 1em plus 0.5em minus 0.4em\relax Springer, 2015, pp. 234--241.

\bibitem{tian2024u}
Y.~Tian, Z.~Tu, H.~Chen, J.~Hu, C.~Xu, and Y.~Wang, ``{U-DiTs}: Downsample tokens in u-shaped diffusion transformers,'' \emph{Advances in Neural Information Processing Systems}, vol.~37, pp. 51\,994--52\,013, 2024.

\bibitem{james2020rlbench}
S.~James, Z.~Ma, D.~R. Arrojo, and A.~J. Davison, ``{RLBench}: The robot learning benchmark \& learning environment,'' \emph{IEEE Robotics and Automation Letters}, vol.~5, no.~2, pp. 3019--3026, 2020.

\bibitem{song2020improved}
Y.~Song and S.~Ermon, ``Improved techniques for training score-based generative models,'' \emph{Advances in Neural Information Processing Systems}, vol.~33, pp. 12\,438--12\,448, 2020.

\bibitem{black2024pi}
K.~Black, N.~Brown, D.~Driess, A.~Esmail, M.~Equi, C.~Finn, N.~Fusai, L.~Groom, K.~Hausman, B.~Ichter \emph{et~al.}, ``$\pi0$: A vision-language-action flow model for general robot control,'' \emph{CoRR}, 2024.

\bibitem{perez2018film}
E.~Perez, F.~Strub, H.~De~Vries, V.~Dumoulin, and A.~Courville, ``{FiLM}: Visual reasoning with a general conditioning layer,'' in \emph{Proceedings of the AAAI Conference on Artificial Intelligence}, vol.~32, no.~1, 2018, pp. 3942--3951.

\bibitem{hoogeboom2023simple}
E.~Hoogeboom, J.~Heek, and T.~Salimans, ``simple diffusion: End-to-end diffusion for high resolution images,'' in \emph{International Conference on Machine Learning}.\hskip 1em plus 0.5em minus 0.4em\relax PMLR, 2023, pp. 13\,213--13\,232.

\bibitem{kaplan2020scaling}
J.~Kaplan, S.~McCandlish, T.~Henighan, T.~B. Brown, B.~Chess, R.~Child, S.~Gray, A.~Radford, J.~Wu, and D.~Amodei, ``Scaling laws for neural language models,'' \emph{arXiv preprint arXiv:2001.08361}, 2020.

\bibitem{yu2023scaling}
T.~Yu, T.~Xiao, A.~Stone, J.~Tompson, A.~Brohan, S.~Wang, J.~Singh, C.~Tan, J.~Peralta, B.~Ichter \emph{et~al.}, ``Scaling robot learning with semantically imagined experience,'' \emph{arXiv preprint arXiv:2302.11550}, 2023.

\bibitem{zhao2023learning}
T.~Z. Zhao, V.~Kumar, S.~Levine, and C.~Finn, ``Learning fine-grained bimanual manipulation with low-cost hardware,'' \emph{arXiv preprint arXiv:2304.13705}, 2023.

\bibitem{carion2020end}
N.~Carion, F.~Massa, G.~Synnaeve, N.~Usunier, A.~Kirillov, and S.~Zagoruyko, ``End-to-end object detection with transformers,'' in \emph{European Conference on Computer Vision}.\hskip 1em plus 0.5em minus 0.4em\relax Springer, 2020, pp. 213--229.

\bibitem{levine2016end}
S.~Levine, C.~Finn, T.~Darrell, and P.~Abbeel, ``End-to-end training of deep visuomotor policies,'' \emph{Journal of Machine Learning Research}, vol.~17, no.~39, pp. 1--40, 2016.

\bibitem{mandlekar2022matters}
A.~Mandlekar, D.~Xu, J.~Wong, S.~Nasiriany, C.~Wang, R.~Kulkarni, L.~Fei-Fei, S.~Savarese, Y.~Zhu, and R.~Mart{\'\i}n-Mart{\'\i}n, ``What matters in learning from offline human demonstrations for robot manipulation,'' in \emph{Conference on Robot Learning}.\hskip 1em plus 0.5em minus 0.4em\relax PMLR, 2022, pp. 1678--1690.

\end{thebibliography}

\end{document}